\documentclass[conference]{IEEEtran}

\usepackage{graphicx}
\usepackage{pgfplots}
\usepackage{float}
\usepackage{caption}
\usepackage{subcaption}
\usepackage[noadjust]{cite}
\usepackage{amsmath}
\usepackage{amssymb}
\usepackage{bm}
\usepackage{amsthm}
\usepackage[numbers]{natbib}
\usepackage[affil-it]{authblk}

\title{Wearable Sensor Data Based Human Activity Recognition using Machine Learning: A new approach}

\author[1]{Nguyen, H.D.} 
\author[2]{Tran, K. P. }
\author[3]{Zeng, X.}
\author[4]{Koehl, L. }
\author[5]{Tartare, G.}

\affil[1]{Faculty of Information Technology, Vietnam National University of Agriculture, Vietnam (e-mail: nhdu@vnua.edu.vn).}
\affil[2]{ENSAIT, GEMTEX – Laboratoire de G\'enie et Mat\'eriaux Textiles, Lille, France(e-mail: kim-phuc.tran@ensait.fr)}
\affil[3]{ENSAIT, GEMTEX – Laboratoire de G\'enie et Mat\'eriaux Textiles, Lille, France (e-mail: xianyi.zeng@ensait.fr)}
\affil[4]{ENSAIT, GEMTEX – Laboratoire de G\'enie et Mat\'eriaux Textiles, Lille, France (e-mail: ludovic.koehl@ensait.fr)}
\affil[5]{ENSAIT, GEMTEX – Laboratoire de G\'enie et Mat\'eriaux Textiles, Lille, France (e-mail: guillaume.tartare@ensait.fr)}

\begin{document}
\maketitle

\textbf{Keyword:}
Human Activity Recognition, Wearable sensor, Ensemble learning method, Machine learning. \\

\begin{abstract}                
Recent years have witnessed the rapid development of human activity recognition (HAR) based on werable sensor data. One can find many practical applications in this area, especially in the field of health care. Many machine learning algorithms such as Decision Trees, Support Vector Machine, Naive Bayes, K-Nearest Neighbor and Multilayer Perceptron are successfully used in HAR. Although these methods are fast and easy for implementation, they still have some limitations due to poor performance in a number of situations. In this paper, we propose a novel method based on the ensemble learning to boost the performance of these machine learning methods for HAR. 
\end{abstract}

\vspace{0.2cm}

\noindent \textbf{ 1. Introduction}
\vspace{0.2cm}

The rapid development of advanced technologies nowadays makes the study of recognizing human activity easier and more effective. Human activity recognition (HAR) is widely used in several fields such as security surveillance, human computer interaction, military, and especially  health care. For instance, the HAR is used for monitoring the activities of elderly people staying in rehabilitation centers for chronic disease management and disease prevention. It  is used to encourage physical exercises for assistive correction of children's sitting posture. It is also applied  in monitoring other behaviours such as abnormal conditions for cardiac patients  and detection for early signs of illness.  More examples of applications of HAR can be seen in \cite{Ong_Ann2014human}. 

 Wearable sensor method refers to the use of smart electronic devices that are integrated into wearable objects or directly with the body in order to measure both biological and physiological sensor signals such as heart rate, blood pressure, body temperature, accelerometers, or other attributes of interest like motion and location. These sensors are communicated with an integration device, like a cellphone, a laptop or a customized embedded system. As a result, the raw signals are sent to an application server for real time monitoring, visualization and analysis. The use of smart mobile phone containing several sensors such as gyroscope, camera, microphone, light, compass, accelerometer, proximity, and GPS can also be very effective for activity recognition (\cite{kwapisz2011activity}). However, the raw data from smart mobile phone is only effective for simple activities but not complex activities (\cite{dernbach2012simple}). Thus, the extra sensors or sensing devices should be used for a better performance of the recognition.

In the literature, various machine learning algorithms have been suggested to handle features extracted from raw signals to identify human activities. These machine learning methods are in general fast and easy to be implemented. However, they only bring satisfying results in a few scenarios because of relying heavily on the heuristic handcrafted feature extraction (\cite{Jindong_Wang2018deep}). Recent years have witnessed the rapid development and application of deep learning, which has also achieved remarkable efficiency in the HAR. Although the deep learning algorithms can automatically learn representative features due to its stacking structure, it has a disadvantage of requiring a large dataset for training model. In fact, numerous data are available  but it is sometimes difficult to access data which are labeled. It is also inappropriate for real-time human activity detection due to the high computation load (\cite{chen2018novel}). This motivates us to apply in this study ensemble algorithm for machine learning based learners  to improve the performance of these algorithms as well as keep the simplicity with implementation. By this new approach, we firstly conduct experiments with several machine learning classifiers such as  Logistic Regression, Multilayer Perceptron, K-Nearest Neighbor, Support Vector Machine, and Random Forest. Then we apply a novel recognition model by using voting algorithm to combine the performance of these algorithms for the HAR. In fact, this algorithm has been suggested in \cite{catal2015use}, leading to impressive results compared to other traditional machine learning methods. In this study, we improve the study in \cite{catal2015use} by using more efficient classifiers as base models. The obtained results show that our proposed method give better performances.

The rest of the paper is organized as follows. In Section 2, we present a brief of related works on HAR. Section 3 describes the methodology using in the study, including the sample generation process, the feature representation, the basic machine learning algorithms and  the proposed voting classifier. The experimental results are shown in Section 4 and some concluding remarks are given in Section 5.

\vspace{0.2cm}
\noindent \textbf{ 2. A brief review of human activity recognition based on wearable sensor data} 
\vspace{0.2cm}

Due to its widely practical applications, the werable sensors based HAR has attracted a large number of studies. A number of machine learning algorithms have been applied to deal with sensors data in the  HAR such as Hidden Markov Models (\cite{lee2011activity}), Support Vector Machines (\cite{anguita2012human}), and K-Nearest Neighbor (\cite{ayu2012comparison}). Other machine learning algorithm like J48 Decision Trees and  Logisitic Regression are also utilized for the HAR based on the accelerometer of smartphone.   An extensive survey on werable sensor-based HAR was carried out by \cite{Lara2013survey}. 

In general, when few labeled data and certain knowledge are required, the machine learning pattern of recognition approaches can give satisfying results. However, there are several limitations of using these methods because of the following arguments:  (1) these regular methods rely heavily on practitioners' experience with heuristic and handcrafted ways to extract interested features; (2) the deep  features are difficult to be learned; and (3) a large amount of well-labeled data to train the mode is required (\cite{Jindong_Wang2018deep}). This is the motivation for the use of deep learning in wearable sensor based HAR recently. \cite{Hammerla_2016deep} used a 5-hidden-layer deep neural network to perform automatic feature learning and classification.  \cite{JianSun_2018} proposed a hybrid deep framework based
on convolution operations, long short-term memory recurrent units, and extreme learning machine classifier. A comprehensive survey on deep learning for sensor-based activity recognition can be seen in \cite{Jindong_Wang2018deep}.

Another method is also widely used for HAR is ensemble learning. This method not only improve significantly the performance of traditional machine learning algorithms and but also avoid a requirement of large dataset for training model of deep learning algorithms. \cite{catal2015use} used a set of classifiers including J48 Decision Trees, Logistic Regression and Multilayer Perceptron, to recognize specifice human activities like walking, jogging, sitting and standing based on accelerometer sensor of a mobile phone. \cite{gandhi2015hybrid} applied a hybrid ensemble classifier that combines the representative algorithms of Instance based learner, Naïve Bayes Tree and Decision Tree Algorithms using voting methodology to $28$ bench mark datasets and compared their method with other machine learning algorithms. A novel ensemble extreme learning machine for HAR using smartphone sensors has been presented in \cite{chen2018novel}.

\vspace{0.2cm}
\noindent \textbf{ 3. Methodology} 
\vspace{0.2cm}

In this Section, we explain the method of generating samples as well as extracting features of data. The individual classifiers and ensemble method using in this study is also presented.

\vspace{0.2cm}
\noindent \textbf{ 3.1 Sample generation process for HAR} 
\vspace{0.2cm}

Generating the samples from raw signasl is a crucial step to perform wearable sensor data based HAR. In general, the raw signals are divided into small parts of the same size, which are called temporal windows, and are used as training and test dataset to define the model. The common process to generate the temporal windows is  semi-non overlapping-window (SNOW). \cite{jordao2018human} pointed out that it is highly biased, i.e, part of the sample's content appears both in the training and testing at the same time. Then, the authors proposed two new methods to handle this bias drawback, including full-non overlapping-window (FNOW) and leave-one-trial- out (LOTO). Although the FNOW can avoid the bias property in SNOW, it has another disadvantage that it provides a less number of  samples compared to the SNOW process. Therefore, the LOTO is proposed to use. In this method, the activities from a trial are initially segmented and then 10-fold cross validation is employed. A figure illustrated the process for the example of 2-folde cross validation is shown in \cite{jordao2018human}. This LOTO process is also applied in  this study.

\vspace{0.2cm}
\noindent \textbf{ 3.2 Feature Representation for HAR} 
\vspace{0.2cm}

Another important step in the process of HAR before applying machine learning algorithms is to extract features from raw data. 
Since the raw signals from sensors are usually noisy, it is necessary to  extract the robust representations, or features from these signals. Several feature representation approaches for HAR have been presented in  \cite{Lara2013survey}. In this study, we focus on a common techniques using for acceleration signals which consists of time- and frequency-domain features. Typical time domain features are mean, standard deviation, variance, root squared mean, and percentiles; while typical frequency domain features include energy, spectral entropy and dominant frequency. (\cite{sani2017learning}). These measures are designed to capture the characteristics of the signal that are useful for distinguishing different classes of activities. 
 Table 1 shows some widely used features from the literature. 
\begin{table}
\center
\begin{tabular}{|c|c|}
\hline
Time Domain Features &Frequency Domain Features\\
\hline
Mean &Dominant frequency\\
Standard deviation &Spectral centroid\\
Inter-quartile range &Energy\\
Kurtosis &Fast Fourier Transform\\
Percentiles  & Discrete Cosine Transform\\
Mean absolute deviation&\\
Entropy&\\
Correlation betwwen axes&\\
\hline
\end{tabular}
\vspace{0.1cm}
\caption{\label{table 1} Hand-crafted features for both time and frequency domains.}
\end{table}

\vspace{0.2cm}
\noindent \textbf{ 3.3 The basic machine learning algorithms} 
\vspace{0.2cm}

\noindent \textit{Logistic Regression}
\vspace{0.1cm}

Logistic regression (LR) is a well-known statistical classification method for modelling a binary response variable, which takes only two possible values. It  simply models probability of  the default class. The LR method is  commonly  used  because  
it  is  easy  with implementation  and  it provides  competitive  results. Although LR is not a classifier, it can still be used to make a classifier or prediction by choosing a cutoff value and classifying inputs with probability greater than the cutoff as one class and less than the cutoff as the other. More detail of the algorithm and its various applications can be seen in \cite{hosmer2013applied}.

\vspace{0.1cm}
\noindent \textit{Multilayer Perceptron}
\vspace{0.1cm}

A multilayer perceptron (MLP) is a classical type of feedforward artificial neural network. It contains one or more layers of neurons. Data is fed to the input layer, there may be one or more hidden layers providing levels of abstraction, and predictions are made on the output layer. The nodes in MLP are fully connected in the sense that each node in one layer connects to every node in the following layer with a certain weight. Based on the amount of error in the output compared to the expected result, the network is trained in the perceptron by changing these connection weights after processing of data. 
 The backpropagation learning technique is applied for the training of the network. The wide applications of MLP can be found in large fields such as speech recognition, image recognition, and machine translation software \cite{wasserman1988neural}.

\vspace{0.1cm}
\noindent \textit{Support Vector Machine}
\vspace{0.1cm}

Support Vector Machine (SVM) is a discriminative classifier belonging to supervised learning models. It constructs a hyperplane (a decision boundary that helps to classify the data points) or set of hyperplanes in a high or infinite dimensional space, which can be used for both classification and regression problem. In practice, there are many possible hyperplanes that could separate the two classes of data points.  The SVM aims to find a hyperplane that has maximum margin, i.e the maximum distance between data points of both classes.  A review on the rule extraction and the main features of the algorithm from SVM can be seen in \cite{barakat2010rule}.

\vspace{0.1cm}
\noindent \textit{Gaussian Naive Bayes}
\vspace{0.1cm}

Gaussian Naive Bayes refers to a naive Bayes classifier as dealing with continuous data in the case that the continuous values associated with each class are distributed following a Gaussian distribution. The naive Bayes classifier is a classification technique based on Bayes' theorem with an independence assumption that the presence of a particular feature in a class is not related to the presence of any other features.  The major advantage of the naive Bayes classifier is
its simplicity and short computational time for training. It can also often outperform some other more sophisticated classification methods (\cite{domingos1997optimality}). 

\vspace{0.1cm}
\noindent \textit{K-Nearest Neighbor}
\vspace{0.1cm}

Being considered as the simplest of all machine learning algorithms, K-nearest neighbour (KNN) is an instance based classifier method. By this algorithm, a new object is classified based on the distance from its $K $ neighbours in the training set and the corresponding weights assigned to the contributions of the neighbors, where the nearer neighbors contribute more to the average than the more distant ones. That is to say, it follows an assumption that similar things are near to each other. The  Euclidean distance is commonly used for continuous variables to calculate the distance in KNN. A review of data classification using this KNN algorithm is given in \cite{kataria2013review}.

\vspace{0.1cm}
\noindent \textit{Random Forest}
\vspace{0.1cm}

Random forest  is an ensemble learning method operated by constructing a multitude of decision trees at training time. Each tree in a random forest learns from a random sample of the data points when training. The samples are drawn with replacement, i.e,  some samples can be used several times in an individual tree. It is trained via the bagging method. Depending on the final task, the output class could be the mode of the classes (for classification) or mean prediction (for regression) of the individual tree. The random forests can overcome the  high risk of overfitting the training data  of the decision trees algorithm. A method of building a forest of uncorrelated trees using a CART (Classification And Regression Tree) and several ingredients forming the basis of the modern practice of random forests has been introduced in \cite{breiman2001random}.

\vspace{0.2cm}
\noindent \textbf{3.4 Proposed voting classifier}
\vspace{0.2cm}

Each machine learning  for classification presented above has their own disadvantages. Ensemble method, a machine learning technique that combines several base models in order to obtain one optimal predictive model, are then proposed in this study.  The main idea of ensemble learning is to aggregate multiple base learners to boost the performance. Several ensemble rules for combining the multiple classification results of different classifiers, including Vote rule, Minimum probability rule, Maximum probability rule, Product of probabilities rule, Median rule,  and Sum Rule have been proposed in \cite{kittler1996combining}. In this study, we apply the voting rule, which is simple but powerful technique, for combining the aforementioned algorithms. Detail of three versions of voting, involving unanimous voting, majority voting and plurality voting, can be found in \cite{wolpert1992stacked}. In unanimous voting, the final decision  is approved by all the base learners. In majority voting, more than 50\% vote is required for final decision and in plurality voting most of the votes decides the final result.  

\begin{figure*}
\centering
\includegraphics[scale=0.5]{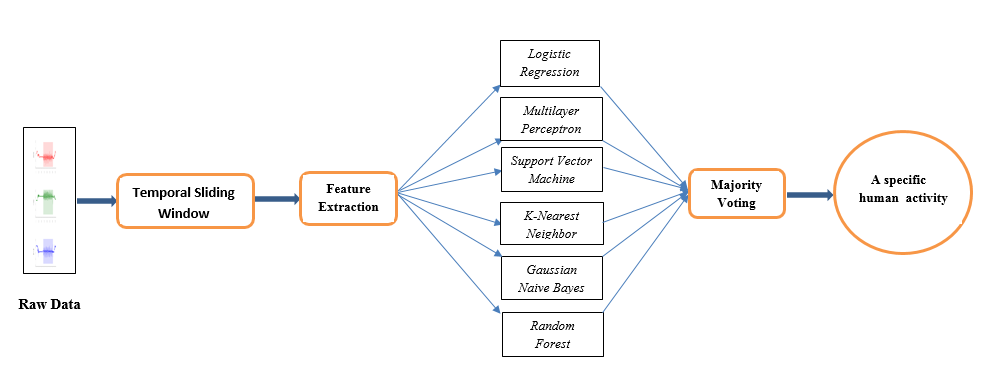}\\
\caption{\small \textit{The proposed framework for HAR using an ensemble algorithm} \label{figure1}}
\end{figure*}

\vspace{0.2cm}
\noindent \textbf{4. Experimental results}
\vspace{0.2cm}

In this Section, we show the experimental results of our proposed method and compare these results with the ones obtained by using the method suggested in \cite{catal2015use}. The two following data sets are considered:
\begin{itemize}
\item MHEALTH: This dataset is based on a new framework for agile development of mobile health applications suggested by \cite{banos2014mhealthdroid}. Four types of signals are provided in this dataset, involving accelerometer signals, gyroscope signals, magnetometer signals and electrocardiogram signals. Similar to \cite{jordao2018human}, we only consider the first three signals in our study.
\item USC-HAD: The USC-HAD, standing for University of Southern California Human Activity Dataset, is developed in \cite{zhang2012usc}. This is specifically
designed to include the most basic and common human activities in daily life from a large and diverse group of human subjects, including 12 activities and 14 subjects. The dataset is available at the website of the authors (see \cite{zhang2012usc}). 
\end{itemize}
\begin{table*}
\centering
\begin{tabular}{|c|c|c|c|c|c|}
\hline
\multicolumn{2}{|c|}{Accuracy}&\multicolumn{2}{|c|}{ Recall}&\multicolumn{2}{|c|}{F-score}\\
\hline
Proposed method&Catal et al.'s method&Proposed method&Catal et al.'s method&Proposed method&Catal et al.'s method\\
\hline
\multicolumn{6}{|c|}{MHELTH}\\
\hline
0.9472 &0.9387&0.9498&0.9423&0.9412&0.9339\\
(\textit{0.9191, 0.9753})&(\textit{0.8941, 0.9834})&(\textit{0.9240, 0.9756})&(\textit{0.9006, 0.9840})&(\textit{0.9100, 0.9725})&(\textit{0.8841, 0.9838})\\
\hline
\multicolumn{6}{|c|}{USCHAD}\\
\hline
0.8690&0.8528&0.8320&0.8174&0.8190&0.8160\\
(\textit{0.8528, 0.8852})&(\textit{0.8335, 0.8721})&(\textit{0.8152, 0.8488})&(\textit{0.7978, 0.8369})&(\textit{0.8027, 0.8354})&(\textit{0.7961, 0.8358})\\
\hline
\end{tabular}
\vspace{0.2cm}
\caption{\small \textit{The experimental results and comparision with the method of \cite{catal2015use}. The italic intervals are the corresponding 90\% confidence intervals.}\label{table2}}
\end{table*}
The first step in experimental setup is temporal sliding window where the samples is split into subwindows and each subwindow is considered as an entire activity. 
We apply the same value of the temporal sliding window size which is $t =5$ seconds as suggested in  \cite{jordao2018human}. Then, the important features from these raw signals (presented in Table \ref{table 1}) are extracted to filter relevant information and to give the input for classifiers. The output of ensemble algorithm is a recognized specific human activity. Figure 1 presents the proposed framework for HAR using ensmble method. Based on the MHEALTH and USC-HAD data set, twelve basic and common activities in people's daily lives is the target to be recognized, including walking forward, walking left, walking right, walking upstairs, walking downstairs, running forward, jumping, sitting, standing, sleeping, elevator up, and elevator down. Table 3 in \cite{zhang2012usc} presents a description for each activity. The performance of proposed method is evaluated by using the following widely used measures:
\begin{itemize}
\item Accuracy = $\frac{TP+TN}{TP+FP+TN+FN}$,
\vspace{0.2cm}
\item Recall = $\frac{TP}{TP+FN}$,
\vspace{0.2cm}
\item F-score = $2\times\frac{\mathrm{Precision}\times \mathrm{Recall}}{\mathrm{Precision}+ \mathrm{Recall}}$.\\
\end{itemize}
where Precision $=TP/(TP+FP)$;  TP (True Positive) is the number of samples  recognized correctly as activities, TN (True Negative) is the number of samples correctly recognized as not activities, FP (False Positive) is the number of samples  incorrectly recognized as activities, and FN (False Negative) is the number of samples incorrectly diagnosed as not activities. 

Our computation was performed on a platform with 2.6
GHz Intel(R) Core(TM) i7 and 32GB of RAM. We perform the experimence on the two datasets mentioned above  using  both our proposed method and the method  suggested in \cite{catal2015use}.  The experimental results are presented in Table \ref{table2}. It can be seen that our method leads to higher values of all the measures being considered. For example, for the MHEALTH data set, our method lead to $Accuracy = 94.72\%$ compared to  $Accuracy = 93.87\%$ from the Catal et al.'s method. Similarly, for the USCHAD data set, $Recall$ is equal to $83.20\%$ corresponding to our propsed method, which is relatively larger than the value $Recall =81.74\%$ corresponding to the Catal et al.'s method.  That is to say, the proposed method in this study outperforms the method used in \cite{catal2015use}. Moreover, the obtained results show that the performance of ensemble learning algorithm can be significantly improved by combining better machine learning algorithms. This should be considered in design a new ensemble method for HAR. 

\vspace{0.2cm}
\noindent \textbf{5. Conclusion and future work}
\vspace{0.2cm}

Activity recognition increasingly plays an important role in many practical applications, especially in health care. Increasing the performance of recognition algorithms is a major concern for researchers. In this paper, we have proposed a new method for wearable sensor data based HAR using the ensemble algorithm. By combining better classifiers, our proposed method improves significantly the previous study (\cite{catal2015use}) in all measures of comparision.

In the future, we would like to address the problem of wearable sensor data based HAR   using Convolutional Neural Networks (CNN) and Long Short-Term Memory (LSTM) networks. The motivation for using these methods is because the CNN offers advantages in selecting good features and the LSTM  has been proven to be good at learning sequential data. 
 \bibliographystyle{plainnat}
\bibliography{SPC_Reference}             
                                                   






\end{document}